\documentclass[a4]{article}

\usepackage{amsmath}
\usepackage{amsfonts}
\usepackage{graphicx}
\usepackage{url}
\usepackage{authblk}
\usepackage{rotating}
\usepackage{float}

\DeclareMathOperator{\MMD}{MMD}

\title{The Kernel Two-Sample Test for Brain Networks}

\author[1,2]{Emanuele Olivetti}
\author[1,2,3]{Sandro Vega-Pons}
\author[1,2]{Paolo Avesani}

\affil[1]{NeuroInformatics Laboratory (NILab), Bruno Kessler
  Foundation, Trento, Italy}

\affil[2]{Center for Mind and Brain Sciences (CIMeC), University of
  Trento, Italy.}

\affil[2]{Pattern Analysis and Computer Vision (PAVIS), Istituto
  Italiano di Tecnologia, Genova, Italy.}

\begin{document}

\maketitle

\begin{abstract}
  In clinical and neuroscientific studies, systematic differences
  between two populations of brain networks are investigated in order
  to characterize mental diseases or processes. Those networks are
  usually represented as graphs built from neuroimaging data and
  studied by means of graph analysis methods. The typical machine
  learning approach to study these brain graphs creates a classifier
  and tests its ability to discriminate the two populations. In
  contrast to this approach, in this work we propose to directly test
  whether two populations of graphs are different or not, by using the
  kernel two-sample test (KTST), without creating the intermediate
  classifier. We claim that, in general, the two approaches provides
  similar results and that the KTST requires much less
  computation. Additionally, in the regime of low sample size, we
  claim that the KTST has lower frequency of Type II error than the
  classification approach. Besides providing algorithmic
  considerations to support these claims, we show strong evidence
  through experiments and one simulation.
\end{abstract}



\section{Introduction}
\label{sec:introduction}

Recent work reported alterations on the structural and functional
connectivity networks in patients with mental diseases like
schizophrenia and
Alzheimer~\cite{supekar2008network,bassett2012altered}. Moreover, in
neuroimaging-based experiments, it has been shown that different
mental states or stimuli can produce alterations on the functional
connectivity
networks~\cite{richiardi2010vector,tagliazucchi2012automatic}.

For these reasons, there is interest in analysis methods that study
systematic differences between populations of networks, usually
represented as graphs. A recent review of the methods available in the
literature~\cite{richiardi2013recent}, describes three main directions
of research in this area: machine learning, statistical hypothesis
testing and network science. In this work we focus on the machine
learning approach and discuss its use within statistical hypothesis
testing.

The use of machine learning on brain networks is a recent and
promising approach with application in both clinical and cognitive
neuroscience~\cite{richiardi2013machine}. Generally, this approach
consists of classifying graphs. Graphs are suitable data structures to
represent brain networks, but they are difficult to be directly
manipulated by machine learning algorithms. Therefore, graph
classification commonly requires an intermediate step in which graphs
are mapped into a feature space, where classifiers can be directly
applied. Such mapping can be done either implicitly, by using graph
kernels~\cite{vishwanathan2010graph}, or explicitly, by using the so
called graph embedding techniques~\cite{fu2013graph}.

Different graph kernels have recently been used for the classification
of neuroimaging data. In the brain decoding literature, we find
applications of the shortest-path kernel~\cite{mokhtari2013decoding},
a custom-designed kernel based on pair-wise node
connectivity~\cite{takerkart2012graph} and the Weisfeiler-Lehman
kernel~\cite{vegapons2013brain, vegapons2014classification}.  This
last one has also been successfully applied to a mild cognitive
impairment (MCI) study~\cite{jie2014topological}.

A popular approach to encode the network information into a graph is
to define a common set of nodes for all graphs, e.g. anatomical
regions of the brain. With this \emph{one-to-one correspondence}
between the nodes across the graphs, the information is stored in the
corresponding weights of the edges, i.e. in the adjacency matrices. In
such a setting, a simple graph embedding technique is obtained by
unfolding the upper triangular part of the adjacency matrix into a
vector. This approach has been used in~\cite{richiardi2011decoding}
for a movie task experiment and in~\cite{tagliazucchi2012automatic}
for the classification of sleep stages. See
Section~\ref{sec:embedding_kernel} for a more detailed description of
graph embeddings and graph kernels.

Classification algorithms, together with graph embeddings or kernels,
are used to study systematic differences among different populations
of brain graphs\cite{richiardi2013recent}. In such application, the
classifier is trained on part of the data, in order to discriminate
the two groups/conditions. Then, on the remaining part of the data,
the classifier is used to quantify how it generalizes the
discrimination to future unseen graphs. This is done by defining a
performance measure, like classification accuracy, that, with respect
of the phenomenon under investigation, is a measure of its
\emph{effect size}. Often, such measures are estimated with resampling
procedures, e.g. cross-validation. Additionally, when using such
measures as a test statistic, it is possible to study the significance
of the phenomenon by means of a statistical procedure, e.g. the
permutation test. Here, we call such test as
\emph{classification-based test} (CBT). The CBT is the core element of
multiple scientific studies~\cite{pereira2009machine}.

During the last decade, the machine learning community developed a
novel way to combine kernel methods with statistical
tests. Specifically, given two populations of objects and a kernel
function to quantify the object's similarity, the \emph{kernel
  two-sample test} (KTST)~\cite{gretton2006kernel,gretton2012kernel}
was proposed to conduct the hypothesis test whether two populations
have the same distribution or not. The KTST is a non-parametric test
based on the \emph{maximum-mean discrepancy} ($\MMD$) test statistic,
a distance function for distributions that can be estimated from
data. The KTST operates a two-sample test between two sets of
arbitrary objects and, differently from CBT, directly quantify
the significance of the effect without creating an intermediate
classifier.

In this work, we propose to use the KTST on brain graphs to directly
test scientific hypotheses. To the best of our knowledge, the KTST has
never been used in this context and the closest works in the
literature are on neuroimaging data not involving
graphs~\cite{olivetti2013kernel,olivetti2014sensor}.

We claim that, in the context of scientific/clinical experiments, the
KTST is an alternative tool to the CBT and that, in general, provides
comparable results.

We also claim that, in case of low sample size, the KTST may have
higher sensitivity than CBT. This is motivated by the fact that the
estimation process of $\MMD$ is simple and deterministic and does not
require the definition of multiple parameters like CBT does. On the
other hand, the test statistic in the CBT requires a more complex and
non-deterministic estimation process, which introduces additional
variability in the result. This occurs because of the competing need
of fitting additional parameters within the classification algorithm,
e.g. the regularization term of support vector machines (SVMs), and of
the non-deterministic process of estimation, i.e. the random
train/test split, especially in the case of
resampling/cross-validation.


In this paper, as support to our claims, we present experiments on
$14$ neuroimaging datasets of graphs, covering different scientific
questions from cognitive studies to clinical investigation, with and
without the node correspondence property. Moreover, we adopt two graph
embeddings and two graph kernels, to show the generality of our
findings. Additionally, we present a simulation study, where we show
graph classification in a simplified setting. With such simulation we
study the probability of Type I and Type II error of KTST and CBT in
case of low sample size. The results of the simulation show the
advantage of KTST over CBT, in terms of lower frequency of Type II
error and equivalent Type I error, and corroborate our findings on
neuroimaging data.

The paper is structured as follows: in Section~\ref{sec:methods} we
introduce the notation, we formally describe graph embeddings and
kernels and we define the CBT and the KTST. In
Section~\ref{sec:materials}, we describe the datasets used in the
experiments. In Section~\ref{sec:experiments}, we provide all the
details and results of the experiments described above and of the
simulation study. In Section~\ref{sec:discussion} and
Section~\ref{sec:conclusions}, we discuss the results and conclude
this work mentioning current limitations and future perspective.


\section{Methods}
\label{sec:methods}
In this section we introduce the notation, some basic concepts and
proceed to explain graphs embeddings, graph kernels and hypothesis
testing. With these ingredients we then formally present the
classification-based test (CBT) and the proposed kernel two-sample
test (KTST).

\subsection{Notation and Basic Concepts}
\label{sec:notation}
Let $G = (V, E, \ell, \omega)$ be a graph, where $V$ is the set of
nodes, $E \subset V\times V$ is the set of edges, $\ell : V
\rightarrow \Sigma$ is a function that assigns a label from an
alphabet $\Sigma$ to each node and $\omega : E \rightarrow \mathbb{R}$
is a function that assigns a real weight value to each edge in the
graph. In this work, the graph $G$ represents the network data of the
brain, e.g. resting state connectivity. Let $\mathcal{G}$ be the space
of all simple, undirected, node-labeled and edge-weighted graphs,
i.e. $G \in \mathcal{G}$.

Let $Y \in \mathcal{Y}$ be a categorical random variable indicating
the population/category of a graph $G$. Here we assume that $Y$ is
binary, e.g. $\mathcal{Y} = \{\mbox{healthy}, \mbox{disease}\}$, and
for notational convenience we indicate $\mathcal{Y} = \{a,b\}$ from
now on.

A neuroimaging experiment over $N$ subjects produces a dataset $D =
\{(g_1,y_1), \ldots (g_N,y_N)\}$ of $N$ graphs drawn i.i.d. from an
unknown probability distribution $P_{\mathcal{G} \times
  \mathcal{Y}}$. In a typical neuroimaging study, $N$ is in the range
$10-200$.

In the terminilogy of two-sample tests, the two samples within $D$ are
$A = \{g : (g,y) \in D, y=a\}$ and $B = \{g : (g,y) \in D, y=b\}$,
where $m=|A|$, $n=|B|$, such that $N = m + n$. From this point of
view, $A$ is sampled from $P_{\mathcal{G}|\mathcal{Y}=a}$ and $B$ from
$P_{\mathcal{G}|\mathcal{Y}=b}$. In the following, for notational
convenience, we call $P_{\mathcal{G}|\mathcal{Y}=a}$ and
$P_{\mathcal{G}|\mathcal{Y}=b}$ as $P_A$ and $P_B$.

\subsection{Graphs: Kernels and Embeddings}
\label{sec:embedding_kernel}

A graph kernel $k: \mathcal{G} \times \mathcal{G} \rightarrow
\mathbb{R}$ is a positive definite kernel function defined on
graphs~\cite{vishwanathan2010graph}, i.e. it is a similarity measure
between graphs, which is symmetric and positive
definite~\cite{hofmann2008kernel}. It is known that, if $k$ is a graph
kernel, there is a mapping $\phi : \mathcal{G} \rightarrow
\mathcal{H}$ from $\mathcal{G}$ to some Hilbert space $\mathcal{H}$,
such that $k(g, g') = \langle \phi(g), \phi(g')\rangle_{\mathcal{H}}$
for all $g, g' \in \mathcal{G}$, where $\langle \cdot, \cdot
\rangle_{\mathcal{H}}$ denotes the inner product in
$\mathcal{H}$. Therefore, graph kernels enable a feature space
representation of graphs in the space $\mathcal{H}$, that allows the
direct application of kernel methods, like support vector machines
(SVMs), on the graph data. Such methods do not require the explicit
representation of graphs in $\mathcal{H}$, but only the evaluation of
the kernel function on pairs of graphs.

During the last years, several graph kernels have been proposed in the
literature. Most of them are based on the ideas of decomposing graphs
into smaller substructures and of building the kernel based on
similarities between those
components~\cite{vishwanathan2010graph}. Following this approach,
there are kernels based on different types of substructures, like
walks~\cite{kashima2003marginalized},
paths~\cite{borgwardt2005shortest} and
trees~\cite{shervashidze2011weisfeiler}. In this paper, we use the
Shortest-path (SP) graph kernel~\cite{borgwardt2005shortest} and the
Weisfeiler-Lehman (WL) graph
kernels~\cite{shervashidze2011weisfeiler}, which have shown good
practical performance and also have a low computational
complexity. The SP kernel computes the similarity between pairs of
graphs based on the number of similar shortest-paths between pairs of
nodes in both graphs. On the other hand, the WL kernel computes the
similarity based on common tree patterns that occur in both graphs.

These graph kernels measure the similarity between graphs based on
differences in their global topology, in terms of paths and trees
patterns. For this reason, they do not directly exploit the
information based on one-to-one correspondence between nodes across
graphs, when available. When the task-related information is in
systematic differences of the graph topology, the graph kernels are
the correct tool to use. Conversely, when there is node
correspondence, e.g. when nodes have anatomical meaning across graphs,
the task-related information may be in the systematic differences in
the weights of the edges and graph kernels may be inadequate to
extract the desired information.

In this case, the explicit embedding of the graphs into a vector space
is an alternative strategy that exploits the one-to-one node
correspondence. Following this idea, the simpler approach is called
Direct Connection Embedding (DCE)~\cite{richiardi2010vector}. It is
basically the computation of a feature vector by unfolding the upper
triangular part of the adjacency matrix of each graph. Another
embedding technique that has been used in this context was introduced
in~\cite{richiardi2010vector} and it is based on the dissimilarity
representation approach for pattern
recognition~\cite{pekalska2005dissimilarity}. The idea is to create a
vector representation of each graph based on its Euclidean distance to
a set of predefined graphs (vectors), called prototypes. In our
experiments, we call this method as Dissimilarity Representation
Embedding (DRE). Both embedding techniques use the information related
to the node correspondence, however they do not directly measure
differences on the global topology of the graph.

In order to use kernel methods, like SVMs or the KTST, a kernel
function must be available. Therefore, graph kernels like SP and WL
can be directly used. On the other hand, embedding techniques like DCE
and DRE can also be used by redefining them as kernel functions. This
can be done by adding a kernel for vector data on the features
obtained by the embedding. More formally, let $e: \mathcal{G}
\rightarrow \mathbb{R}^d$ be a graph embedding function, e.g. DCE, we
can define the corresponding kernel function $k_e: \mathcal{G} \times
\mathcal{G} \rightarrow \mathbb{R}$ as
$$k_e (g, g') = k_v (e(g), e(g'))$$ 
where $k_v$ is a kernel function for vector data, like the linear or
Gaussian (RBF) kernels.


\subsection{Hypothesis Testing}
\label{sec:ht}
There are different schools of thought about testing hypotheses, which
define different testing procedures. Here we adopt the Frequentist
framework because it is the main one in the experimental neuroscience
field. Moreover it is the one adopted both by CBT and KTST. Within the
Frequentist framework, the two main schools of thought are those
associated with Fisher~\cite{fisher1935design} and
Neyman-Pearson~\cite{neyman1933problem}. In our experiments, described
in Section~\ref{sec:experiments}, we adopted both views for different
purposes. For this reason, here we briefly report the two procedures
and their aim.

\subsubsection{Fisher significance testing}
\label{sec:fisher}
the procedure for testing hypotheses defined by
Fisher~\cite{fisher1935design}, is based on the definition of the
hypothesis to disprove with the experiment, i.e. the \emph{null
  hypothesis} $H_0$, and by quantifying the evidence in the data
against it. It comprises the following steps:
\begin{enumerate}
\item Set up the null hypothesis $H_0$ to disprove.
\item Define an appropriate test statistic $T$, which is a function
  that, given the collected data, summarizes them in a real number.
\item Compute $p(T| H_0)$, i.e. the distribution of $T$ when $H_0$ is
  true, for example with resampling techniques.
\item Run the experiment, collect the data and compute $T^*$ as the
  value of the test statistic for the observed data.
\item Compute the $p\mbox{-value} = p(T \ge T^*| H_0)$, as the
  probability of getting an equal or more extreme value of $T^*$ when
  $H_0$ is true.
\item Report the $p$-value as a quantification of the evidence against
  $H_0$.
\end{enumerate}
The interpretation of the $p$-value as a measure of evidence against
$H_0$ has been subject to debate in the
literature~\cite{sellke2001calibration}. Nevertheless, it is common
practice to define a threshold $\theta$ for the $p$-value, below which
the result is considered \emph{significant}. Typical values for are
$\theta = 0.05$ or $\theta = 0.01$.

\subsubsection{Neyman-Pearson hypothesis testing}
\label{sec:np}
the procedure to test hypotheses defined by Neyman and
Pearson~\cite{neyman1933problem} is based on defining two alternative
hypotheses and to decide which one to accept, by characterizing the
test through the probability of Type I ($\alpha$) and Type II
($\beta$) error and the sample size $N$. These are the steps:
\begin{enumerate}
\item Set up two complementary hypotheses, $H_0$ (null) and $H_1$
  (alternative).
\item Define an appropriate test statistic $T$.
\item Trade-off $\alpha = p(\mbox{reject } H_0| H_0 \mbox{ is
    true})$, $\beta = p(\mbox{reject } H_1| H_1 \mbox{ is true})$ and
  the sample size $N$, to fit the goals of the experiment.
\item Compute the rejection region(s) $\mathcal{R}$ for $T$, where
  $H_1$ is accepted and $H_0$ rejected.
\item Run the experiment, collect the data and compute $T^*$ as the
  value of the test statistic for the observed data.
\item Reject $H_0$ and accept $H_1$ if $T^* \in \mathcal{R}$. Or
  viceversa, if $T^* \notin \mathcal{R}$.
\end{enumerate}
A test is called \emph{consistent} if $\beta \rightarrow 0$ as $n
\rightarrow \infty$ whenever $H_0$ is false.

Common steps for the Fisher and Neyman-Pearson approaches are the
definition of the test statistic and the derivation of its null
distribution. In the following we describe this two steps for the CBT
and the KTST.

\subsection{Classification-Based Test}
\label{sec:cbt}
A classifier $c \in \mathcal{C}$ is a function $c:\mathcal{G} \mapsto
\mathcal{Y}$ that returns the predicted class label given a
graph. Typically classifiers need training to be instantiated. This
requires that a portion of the dataset $D$, called train set
$D_{{train}}$, is used to fit the classifier to the data.

In the classification-based test (CBT), i.e. when testing whether a
classifier is able to discriminate the two classes $\{a,b\}$, the null
hypothesis $H_0$ is that $c$ predicts at chance-level and the
alternative hypothesis $H_1$ is that $c$ predicts better than
chance-level\footnote{In this work we do not discuss
  \emph{antilearning}~\cite{kowalczyk2005analysis}, the rare event
  where the classifier performs systematically worse than
  chance.}. The most common measure to quantify the ability of
classifier to discriminate the classes is the generalization error
$\epsilon = E_{\mathcal{G} \times \mathcal{Y}}[I(Y,c(G))]$, where $I$
is the indicator function. The standard unbiased estimator of
$\epsilon$ is the \emph{error rate} $\hat{\epsilon} =
\frac{1}{D_{{test}}} \sum_{(g,y) \in D_{{test}}}I(y,c(g))$, where
$D_{{test}} = D \setminus D_{{train}}$. Here we adopt the
complementary measure of the error rate, i.e. accuracy $acc = 1 -
\hat{\epsilon}$, which is more common in neuroscience applications.

When $D_{test}$ is imbalanced, i.e. when $n$ and $m$ considerably
differ, the interpretation of accuracy as a measure of discrimination,
can problematic~\cite{brodersen2010balanced,olivetti2012induction}. An
alternative measure that reduces the impact of this problem is
\emph{balanced accuracy}~\cite{brodersen2010balanced}, i.e. the
average per-class accuracy $acc_B = \frac{1}{2}\left( \frac{TP}{m} +
  \frac{TN}{n} \right)$, where $TP$, $TN$ are the true negatives and
true positives, i.e. the correctly classified examples of class $A$
and $B$, respectively. Notice that, for $acc_B$, the chance level is
always $0.5$, irrespective of how imabalanced the data are.

The estimation of performance measure like $acc$ or $acc_{B}$ may have
high variability for small $N$. Moreover, the split of $D$ in
$D_{{train}}$ and $D_{{test}}$ is stochastic, adding more variability
to the estimate. In order to reduce this issue, it is common to adopt
a resampling technique, the most common in this context being
$\kappa$-folds cross-validation (CV). In CV, $D$ is randomly split in
$\kappa$ non-overlapping parts and, iteratively, one is used as test
set and the remaining parts as train set. The cross-validated balanced
accuracy, $acc_{CV}$, is then the average $acc_B$ across the folds.

In practical cases, most classifiers have additional parameters,
called hyperparameters, that must be set before training, e.g. the
regularization term of SVMs. For this reason, part of the data need to
be used to assess such parameters. In order to avoid circularity, care
has to be taken in selecting the portion of the data for this
step. The standard process for unbiased estimation of hyperparameters,
training and estimation of the error rate requires a nested CV scheme,
as described in~\cite{olivetti2010braindecoding}. Notice that
hyperparameter estimation, training and error rate estimation compete
in the exclusive use of the data, because the more the data is used
for one, the less remain for the others. The effect of this
competition is further uncertainty in the estimates, which may reach
critical levels for small $N$.

In the CBT, the usual test statistic $T$ is $acc_{CV}$. In practical
cases, the null distribution of $acc_{CV}$ is estimated through
resampling, specifically through permutations of the class labels in
$D$. In this work we adopt the standard Monte Carlo approximation of
the permutation approach in which, for $M$ iterations, the examples
are randomly assigned the group $A$ or $B$ and the permuted $acc_{CV}$
is computed. The null distribution of $acc_{CV}$ is then approximated
by the obtained M values. The approximated $p$-value of the observed
(unpermuted) $acc_{CV}^*$ is then the fraction of the $M$ values
greater then $acc_{CV}^*$.

Given $\kappa$ folds, $v$ hyperparameter values and $M$ permutations,
the computational cost of CBT is dominated by the $M v \kappa^2$
trainings and testings for the estimation of the null distribution,
because of the nested cross-validation for each permutation and
because in the inner loop all hyperparameters are attempted. Notice
that, usually, training and testing are computationally expensive and
even for small datasets CBT may require a large amount of time to be
computed.

\subsection{Kernel Two-Sample Test}
\label{sec:ktst}
A two-sample problem compares samples from two probability
distributions $P_A$ and $P_B$. A two-sample test is an hypothesis test
where the null hypothesis $H_0 : P_A = P_B$ and the alternative
hypothesis $H_1 : P_A \neq P_B$ are tested given the data.

Two sample tests can be parametric, like the Student's $t$-test in one
dimension or the Hotelling $T$-test in higher dimensions, where $P_A$
and $P_B$ are Gaussians. These tests do not directly apply to graphs
because they require real values or vectors. Moreover, in the
high-dimensional setting, e.g. the one obtained through graph
embedding, the $T$-test performs poorly, as explained
in~\cite{bai1996effect}.

To address the high-dimesional setting or more general topological
spaces, some non-parametric two-sample tests have been proposed in the
literature. For a brief review see~\cite{gretton2012kernel} Section
3.3. Among them, the most popular one is the kernel two-sample test
(KTST)~\cite{gretton2006kernel,gretton2012kernel} which is based on
the \emph{maximum mean discrepancy} ($\MMD$) test statistic:
\begin{equation}
  \label{eq:mmd}
  \MMD[P_A,P_B] = \max_{||f||_{\mathcal{H}} \leq 1} \left( E_{P_A}[f(x_A)]
    - E_{P_B}[f(x_B)]\right)
\end{equation}
where the function $f$ is from the unit ball in a reproducing kernel
Hilbert space $\mathcal{H}$ and $x_A, x_B \in \mathcal{X}$ are objects
from a generic topological space $\mathcal{X}$, sampled according to
$P_A$ or $P_B$. Notice that $x$ is not necessarily a vector but can be
any object, like a graph, for which $\mathcal{H}$ is defined. As
mentioned in Section~\ref{sec:embedding_kernel}, $\mathcal{H}$ can be
defined through a kernel function $k(x,x'): \mathcal{X} \times
\mathcal{X} \mapsto \mathbb{R}$. In~\cite{gretton2012kernel}, an
interesting property of $\MMD$ is proved: for some families of
kernels called \emph{characteristic} kernels, like the Gaussian and
Laplacian kernels~\cite{fukumizu2008kernel},
\begin{equation}
  \label{eq:mmd_injective}
\MMD[P_A,P_B] = 0 \mbox{\quad if and only if \quad} P_A = P_B. \nonumber
\end{equation}
when $k$ is bounded. This result means that the KTST is consistent for
such kernels.
In~\cite{gretton2012kernel} the following relationship between $\MMD$
and the given kernel is derived:
\begin{align}
  \label{eq:mmd2_kernel}
  \MMD^2[P_A,P_B] = \ &  E_{x_A,x'_A}[k(x_A,x'_A)] \nonumber + \\
  & - 2 E_{x_A,x_B}[k(x_A,x_B)] + \nonumber \\
  & + E_{x_B,x'_B}[k(x_B,x'_B)]
\end{align}

In practical cases, we do not have access to $P_A$ and $P_B$, but just
to the samples $A$ and $B$. In~\cite{gretton2012kernel}, an unbiased
estimate of $\MMD^2$ is derived
\begin{align}
  \label{eq:mmd2u}
  \MMD^2_u(A, B) = \ & \frac{1}{m(m-1)}\sum_{i \neq j}k(x^A_i,x^A_j) +
  \nonumber \\
   & - \frac{2}{mn} \sum_{i,j}k(x^A_i,x^B_j) + \nonumber \\
   & + \frac{1}{n(n-1)}\sum_{i \neq j}k(x^B_i,x^B_j)
\end{align}

Notice that $\MMD$ and $\MMD^2_u$ are not absolute measures of the
differences between two distributions. For example, if we use two
different kernels on the same data / distributions, we cannot
numerically compare the two $\MMD^2_u$. The same issue occurs when
computing $\MMD$ from two different problems, pertaining to different
domains and distributions. From this point of view, in the context of
experimental neuroscience described here, $\MMD$ cannot be used as a
measure of the \emph{effect size} of the phenomenon under
investigation.

The null distribution of $\MMD^2_u$ can be estimated in different
ways, as described in~\cite{gretton2012kernel}. In case of small
samples, which is typical of the neuroimaging domain, a resampling
approach is suggested, which here we implement as permutation test. In
this work we adopt the standard approximation of the permutation test,
as described for the CBT in Section~\ref{sec:cbt}. This requires
computing $\MMD^2_u$ for each of the $M$ permutations, for some large
$M$, e.g. $M = 10000$,

Given $N$ subjects/examples, $M$ permutations and assuming that the
kernel matrix $K = [k(x_i,x_j]_{ij = 1 \ldots N}$ is precomputed in
advance, the computational cost of the KTST is approximately of $M
N^2$ sums.


\section{Materials}
\label{sec:materials}
In this section we describe the datasets used in the experiments of
Section~\ref{sec:experiments}.

\subsection{1000 Functional Connectomes Dataset}
\label{sec:1000fc}
The first dataset corresponds to the connectivity matrices computed
from the resting state fMRI data acquired under the 1000 Functional
Connectome
Project\footnote{\url{http://www.nitrc.org/projects/fcon_1000}}. This
dataset is publicly available~\cite{biswal2010toward}, and
particularly, the functional connectivity matrices can be downloaded
from the USC Multimodal Connectivity
Database~\cite{brown2012multimodal} under the name
\emph{1000\_Functional\_Connectomes}\footnote{\url{http://umcd.humanconnectomeproject.org/umcd/default/browse_studies}}.

We use this dataset for a gender classification problem,
i.e. $\mathcal{Y}$ = $\{$\mbox{male}, \mbox{female}$\}$, motivated by
a similar experiment in~\cite{casanova2012combining}. In this dataset,
all brain data was motion-corrected and normalized to a standard
template where $177$ brain regions were defined. Therefore, all graphs
are composed of $|V| = 177$ nodes, and it is possible to establish the
correspondence between nodes representing the same brain region.

This dataset is a collection of fMRI datasets recorded at different
locations all over the world. In our experiments, we grouped the data
according their location and each location-specific dataset was used
independently, to avoid batch effects due to different MRI scanners.
Additionally, we discarded the locations-specific datasets with too
few subjects, i.e. when either $m < 10$ or $n < 10$. In such cases, it
is expected that both CBT and KTST are not able to reject $H_0$ when
it is indeed false, just for lack of data. By discarding those
excessively small datasets, we avoided to include cases where the
agreement between CBT and KTST was granted because of the sample size
and not because of the actual information content in the data. In
other words, we carefully prevented to artificially inflate the
claimed agreement between CBT and KTST.

In this way, this large dataset is transformed into $12$ smaller
datasets, each one containing the data of a specific location. The
name of the remaining locations and their corresponding number of
subjects per class is reported in Table~\ref{tab:results_fmri}.

\subsection{Schizophrenia Dataset}
\label{sec:schizophrenia}
We used the functional connectivity dataset released with the MLSP
2014 Schizophrenia Classification
Challenge~\footnote{\url{http://www.kaggle.com/c/mlsp-2014-mri/data}}. This
data is partially described in~\cite{cetin2014thalamus}. The dataset
is composed by $86$ functional connectivity matrices, $m = 46$
belonging to the control class and $n = 40$ to the schizophrenia
class. The brain data of all subjects was parcelled into $28$ regions
and therefore all graphs have $|V| = 28$ nodes, and the node
correspondence could be established. The weights on the edges
correspond to the correlation between time series of every pair of
brain regions.

\subsection{Contextual Disorder Dataset}
\label{sec:contextual_disorder}
The third dataset corresponds to a listening task experiment with fMRI
data recently proposed in~\cite{andric2015global}. In this study, $19$
healthy participants were presented with two types of sequences of
auditory stimuli: $\mathcal{Y}$ = $\{$\mbox{ordered},
\mbox{disordered}$\}$, as well as two other conditions not discussed
here. Each sequence, of $150s$, was presented once to each
participant, therefore the dataset is composed of $38$
examples. Connectivity graphs were computed by following the approach
presented in~\cite{vegapons2014classification}. First, a hierarchical
clustering algorithm was independently applied on each participant's
data in order to define brain regions, i.e. the nodes the graph. Then,
correlation values between the average timeseries of the pairs of
regions were used to define the edge weights.

In this dataset the data of different participants were processed
independently. Therefore, graphs computed from different participants
had different number of nodes and there was no anatomical
correspondence between any pair of nodes across the graphs.

\subsection{Simulated Data}
\label{sec:simulated_data}
We generated multiple simulated datasets each consisting of $m = n =
20$ star graphs with $d+1$ nodes and $d$ edges. We assumed node
correspondence and defined $P_A$ and $P_B$ as $d$-dimensional
multivariate normal distributions of weights of the edges. $P_A$ and
$P_B$ had same covariance but different mean: $\mu_A=(0,\ldots,0) \in
\mathbb{R}^d$, $\mu_B=(\delta,\ldots,\delta) \in \mathbb{R}^d$, thus
$\delta$ was the effect size. We generated $1000$ datasets with $d=5$
and $\delta \in \{0.0, 0.25, 0.5, 0.75, 1.0\}$, simulating both cases,
i.e when $H_0: P_A = P_B$ ($\delta = 0$) is true and when $H_0$ is
false ($\delta \in \{0.25, 0.5, 0.75, 1.0\}$), with different effect
sizes.


\section{Experiments}
\label{sec:experiments}
With a set of experiments, we investigated the degree of agreement
between KTST and CBT, to see whether KTST is a viable alternative to
CBT. In order to do that, in this section we show the results of CBT
and KTST on 14 datasets, spanning different scientific questions
and using two graph embeddings (when possible) and two graph
kernels. This section is concluded with a simulation study where, on a
simplified example, the probability of Type I and Type II error is
quantified both for CBT and KTST, for sample sizes analogous to those
of typical neuroimaging experiments.

In all experiments, in order to avoid biases and to keep the
experimental conditions most similar between CBT and KTST, we used
SVMs as classifier\footnote{For model selection, the regularization
  parameter of SVM was optimized over $v = 25$ values ranging from
  $10^{-5}$ to $10^5$, equally spaced in log-scale.} so to use the
same kernel matrix for CBT and KTST. The kernel matrices related to
graph embeddings were based on the Gaussian kernel, as mentioned in
Section~\ref{sec:embedding_kernel}, using the median value of the
distances between the graphs, as the $\sigma$ parameter. This is a
standard heuristic in case of low sample size, because it avoids
spending class-labeled data to fit $\sigma$. All kernel matrices of
all experiments were pre-computed in advance.

The CBT test statistic was the $5$-fold cross-validated balanced
accuracy $acc_{CV}$ in all cases. The null distributions of $acc_{CV}$
and $\MMD^2_ u$ were approximated with $M=10000$ permutations. As
noted in Section~\ref{sec:cbt}, the actual value of $acc_{CV}$ depends
on the random train/test split during the (nested) cross-validation
process, which introduces variability in the result. In the following,
in Table~\ref{tab:results_fmri} and
Figure~\ref{fig:p_values_KTST_vs_CBT}, we report the \emph{median}
$acc_{CV}$ and its related $p$-value after $100$ repetitions of the
estimation process. Differently, the $\MMD^2_u$ value is deterministic
and has no such variability.

\begin{figure}
  \centering
  \includegraphics[width=9cm]{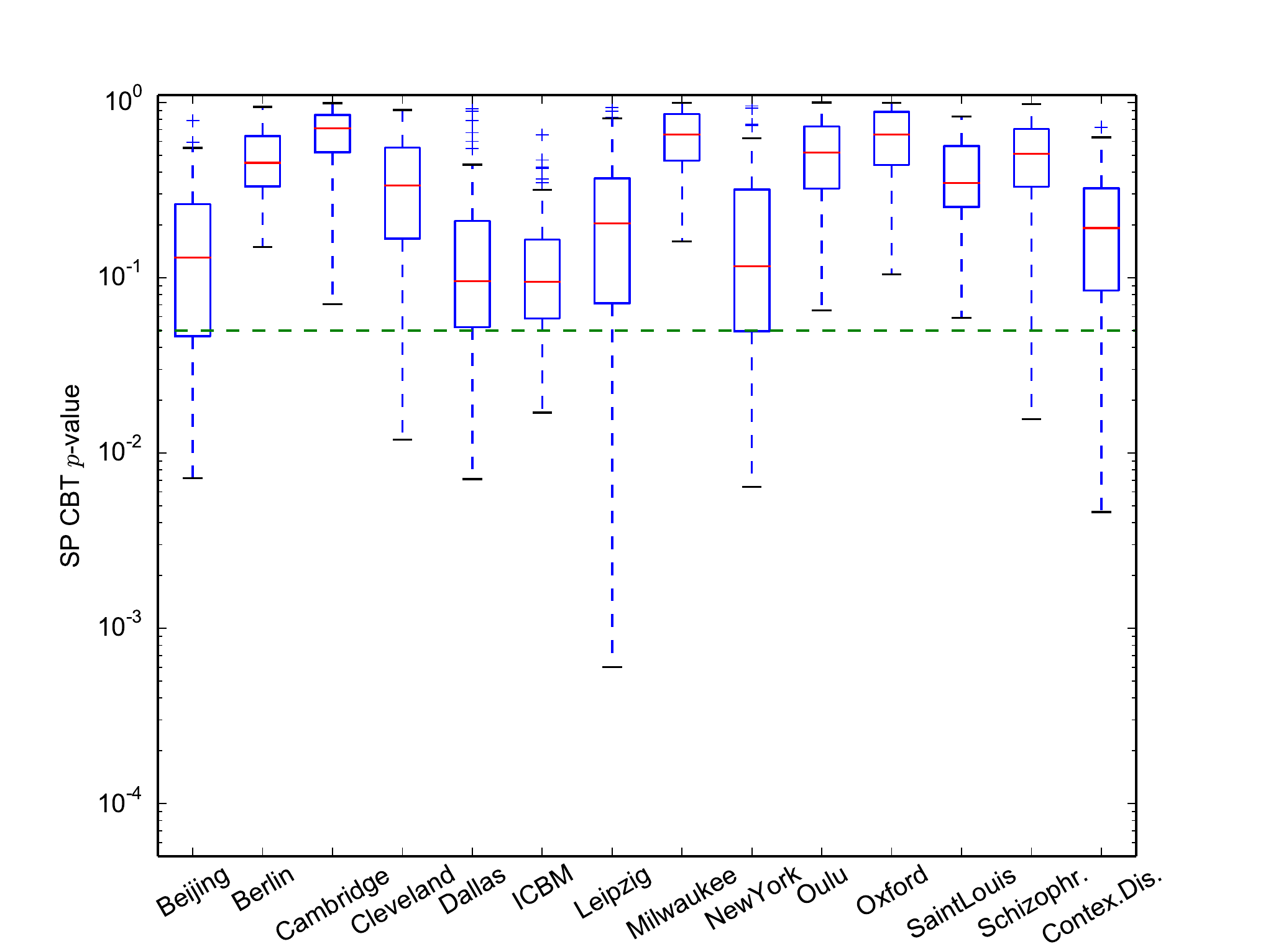}
  \caption{For each dataset of the experiments, the boxplots represent
    the variability of the $p$-values of CBT with SP kernel across
    multiple runs. The horizontal dashed line is the threshold for
    significance $\theta=0.05$.}
  \label{fig:p_values_CBT_SP_boxplot}
\end{figure}

In Figure~\ref{fig:p_values_CBT_DCE_boxplot},
\ref{fig:p_values_CBT_DRE_boxplot}, \ref{fig:p_values_CBT_WL_boxplot}
and~\ref{fig:p_values_CBT_SP_boxplot} we explicitly show the
variability of the $p$-values associated to the multiple estimates of
$acc_{CV}$ through boxplots, because this is an important element for
the discussion in Section~\ref{sec:discussion}.

We report that, in each single test, the time to compute CBT and KTST
greatly differed, according to what was mentioned in
Section~\ref{sec:cbt} and Section~\ref{sec:ktst}. In the typical case
of $m = n = 30$, CBT required hours of computation on a modern
$4$-cores computer, while KTST required just a few seconds.

\subsection{Agreement between CBT and KTST}
\label{sec:agreement}
We carried out the CBT and KTST on the brain graphs from 14 datasets,
coming from three different scientific studies: the 1000 Functional
connectome, a schizophrenia study and a cognitive neuroscience study
about processing ordered and unordered sequences of auditory
stimuli. In Section~\ref{sec:materials} are reported the details of
each study. Here we mention that, in the first two studies, the data
is preprocessed in order to provide node correspondence, which is
ideal for graph embeddings, while in the third study there is no node
correspondence, which allows the use of only graph kernels.

From the first study, i.e. the 1000 Functional Connectome, we analyzed
12 dataset from different recording locations in order to investigate
their relation to the gender of the subject. In these dataset, the
size of the subgroups, i.e. male and female, ranges from 10 to
123. The second study, about schizophrenia, comprises $40$ patients
and $46$ healthy controls. The third study, about ordered and
unordered auditory stimuli, provides data from $19$ subjects, each
with one recording session for both categories of stimulus.

In Table~\ref{tab:results_fmri}, for each dataset, we report the size of
each dataset ($n$ and $m$), the balanced accuracy $acc_{CV}$ and the
$p$-values of CBT and KTST for two different graph embeddings,
i.e. DCE and DRE, and for two different graph kernels, i.e. WL and SP,
as explained in Section~\ref{sec:embedding_kernel}.

\begin{figure}[t]
  \centering
  \includegraphics[width=9cm]{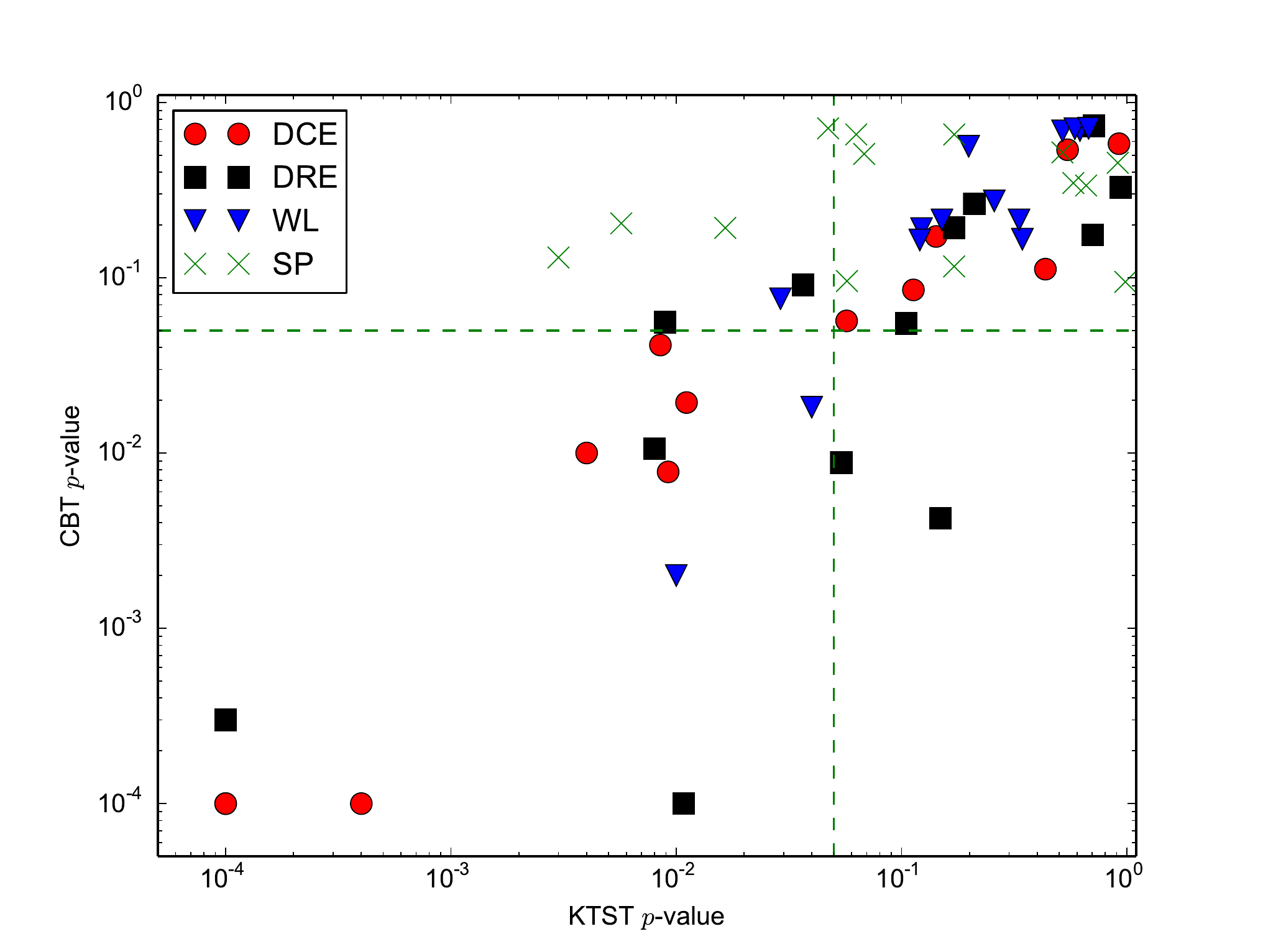}
  \caption{Results of Table~\ref{tab:results_fmri} represented as the
    $p$-value of KTST vs. the p-value of CBT. Each point corresponds
    to a dataset and the shape/color represents the embedding or
    kernel used. The horizontal dashed lines are the thresholds for
    significance $\theta=0.05$.}
  \label{fig:p_values_KTST_vs_CBT}
\end{figure}

In Figure~\ref{fig:p_values_KTST_vs_CBT} we plot the results of
Table~\ref{tab:results_fmri} in log-log scale, to show the agreement
between CBT and KTST. There, each location is represented as point
with coordinates given by the $p$-value of CBT and $p$-value of KTST,
and different colors are used for different embeddings or kernels.

\begin{figure}
  \centering
  \includegraphics[width=9cm]{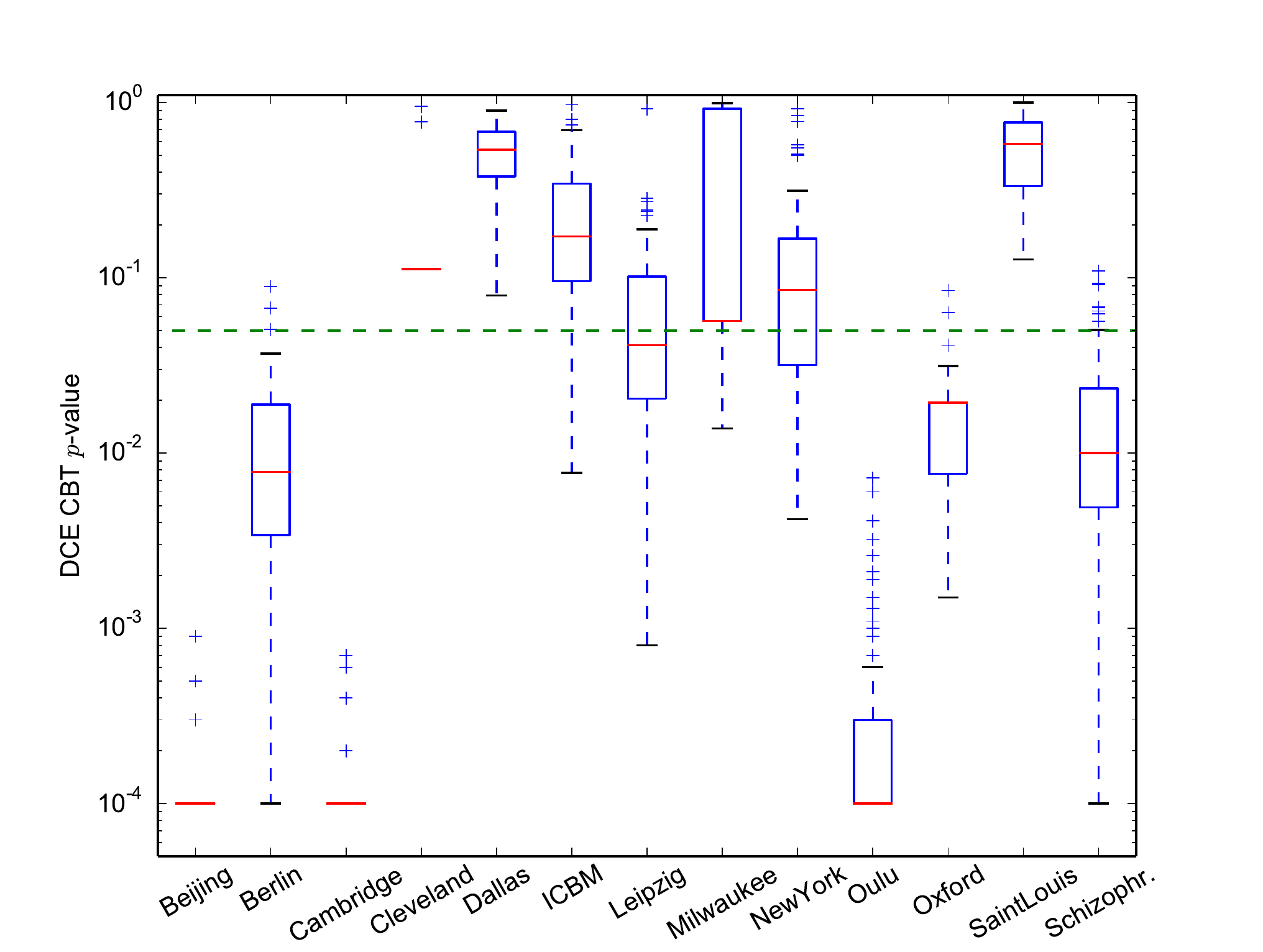}
  \caption{For each dataset of the experiments, the boxplots represent
    the variability of the $p$-values of CBT with DCE across multiple
    runs. The horizontal dashed line is the threshold for
    significance $\theta=0.05$.}
  \label{fig:p_values_CBT_DCE_boxplot}
\end{figure}

\begin{figure}
  \centering
  \includegraphics[width=9cm]{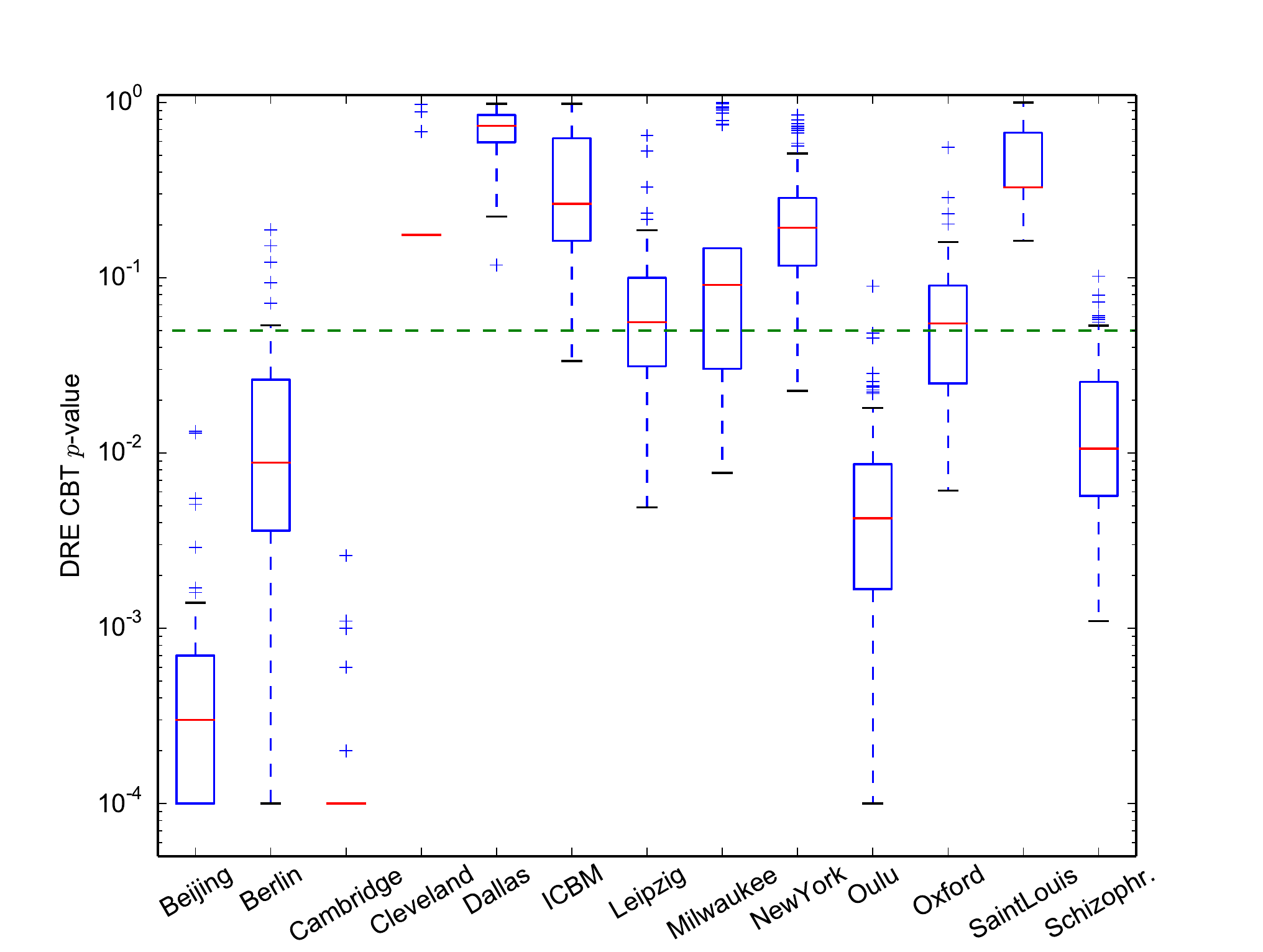}
  \caption{For each dataset of the experiments, the boxplots represent
    the variability of the $p$-values of CBT with DRE across multiple
    runs. The horizontal dashed line is the threshold for
    significance $\theta=0.05$.}
  \label{fig:p_values_CBT_DRE_boxplot}
\end{figure}

\begin{figure}
  \centering
  \includegraphics[width=9cm]{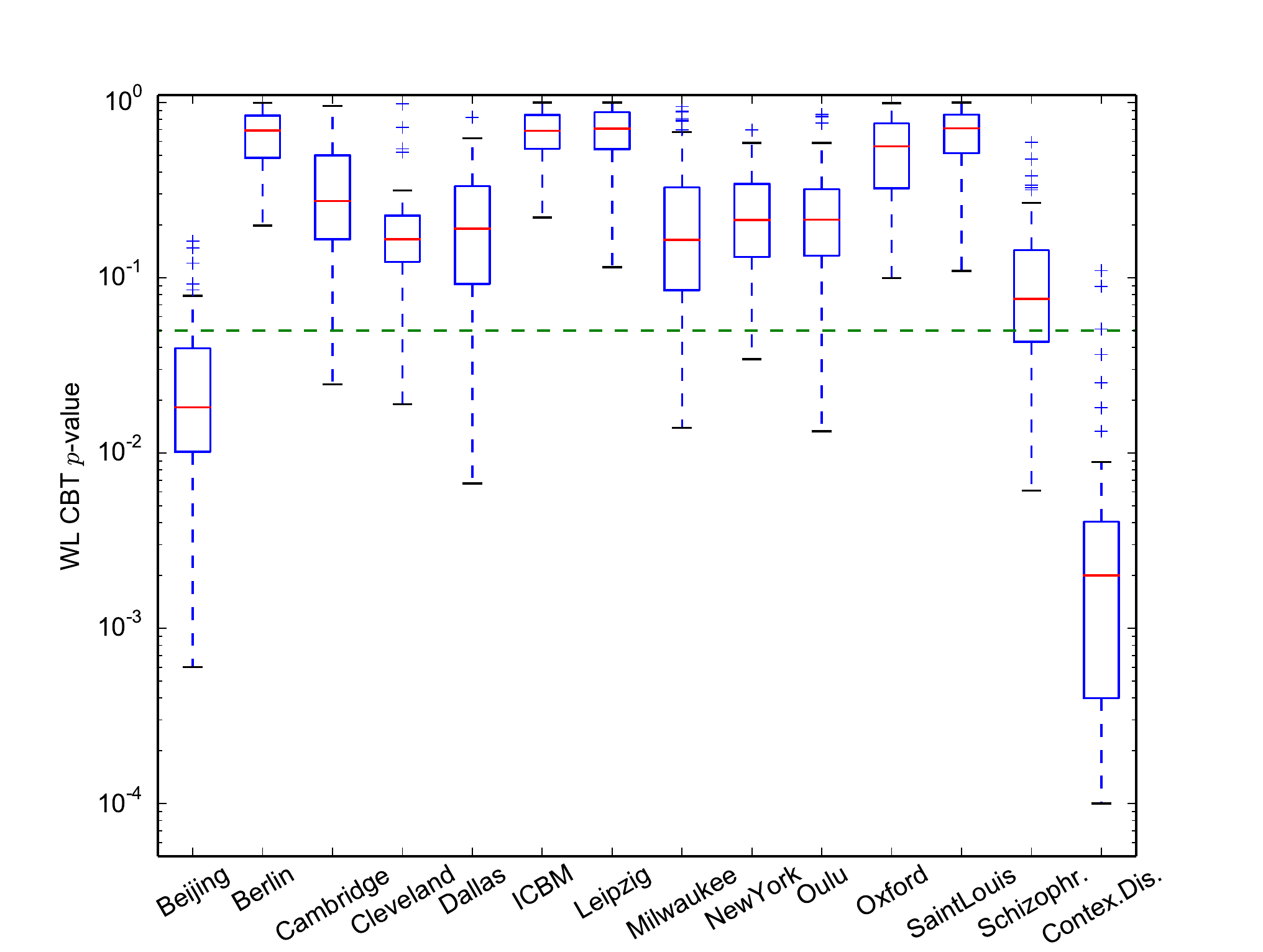}
  \caption{For each dataset of the experiments, the boxplots represent
    the variability of the $p$-values of CBT with the WL kernel across
    multiple runs. The horizontal dashed line is the threshold for
    significance $\theta=0.05$.}
  \label{fig:p_values_CBT_WL_boxplot}
\end{figure}

As mentioned before, in Figure~\ref{fig:p_values_CBT_DCE_boxplot},
\ref{fig:p_values_CBT_DRE_boxplot}
and~\ref{fig:p_values_CBT_WL_boxplot} we explicitly show the
variability of the $p$-values associated to the multiple estimates of
$acc_{CV}$, through boxplots, for each of the 14 datasets. We do not
report equivalent boxplots for the $p$-values of KTST because
$\MMD^2_u$ is deterministic and the only (negligible) source of
variability was the approximation of the null distribution. With
$M=10000$ iterations, those $p$-values were always stable to the
reported 3rd decimal place.

\begin{sidewaystable}[t]
  \centering
  \caption{Results of experiments on all FMRI datasets. Name/Location
    and size ($n$, $m$) are reported in the first three columns. Across
    two graph 
    embeddings (DCE, DRE) and two graph kernels (WL, SP), the median
    $acc_{cv}$ and the
    related $p$-value are reported for CBT, while just the $p$-value
    for KTST. In bold the cases of
    disagreement in the $p$-value of CBT vs. KTST,
    assuming a standard threshold for significance $\theta=0.05$.}
  \label{tab:results_fmri}
  \begin{tabular}{l | c | c | c | c | c | c | c | c | c | c | c | c | c | c }
\multicolumn{3}{c}{} & \multicolumn{3}{| c |}{DCE} & \multicolumn{3}{| c |}{DRE} & \multicolumn{3}{| c }{WL kernel}  & \multicolumn{3}{| c }{SP kernel} \\
\hline
\multicolumn{3}{c}{} & \multicolumn{2}{| c |}{CBT-SVM} & KTST & \multicolumn{2}{| c |}{CBT-SVM} & KTST & \multicolumn{2}{| c |}{CBT-SVM} & KTST & \multicolumn{2}{| c |}{CBT-SVM} & KTST \\
\hline
Dataset & $m$  & $n$  & $acc_{CV}$ & $p$-val & $p$-val & $acc_{CV}$ & $p$-val & $p$-val & $acc_{CV}$ & $p$-val & $p$-val & $acc_{CV}$ & $p$-val & $p$-val \\
\hline

Beijing      & 109 &  65 & 0.655 & 0.000 & 0.000 & 0.637 & 0.000          & 0.000          & 0.590 & 0.018 & 0.040 & 0.532 & \textbf{0.130} & \textbf{0.003} \\
Berlin       &  13 &  13 & 0.767 & 0.008 & 0.009 & 0.767 & \textbf{0.009} & \textbf{0.054} & 0.433 & 0.692 & 0.619 & 0.500 & 0.453 & 0.911 \\
Cambridge    & 123 &  75 & 0.660 & 0.000 & 0.000 & 0.647 & 0.000          & 0.011          & 0.513 & 0.274 & 0.258 & 0.484 & \textbf{0.711} & \textbf{0.047} \\ 
Cleveland    &  20 &  11 & 0.500 & 0.112 & 0.435 & 0.500 & 0.176          & 0.705          & 0.550 & 0.166 & 0.343 & 0.521 & 0.336 & 0.659 \\
Dallas       &  12 &  12 & 0.483 & 0.537 & 0.545 & 0.417 & 0.735          & 0.714          & 0.600 & 0.191 & 0.123 & 0.617 & 0.096 & 0.057 \\
ICBM         &  29 &  22 & 0.538 & 0.172 & 0.142 & 0.518 & 0.264          & 0.210          & 0.466 & 0.688 & 0.518 & 0.580 & 0.095 & 0.984 \\
Leipzig      &  21 &  16 & 0.617 & 0.041 & 0.009 & 0.611 & \textbf{0.056} & \textbf{0.009} & 0.458 & 0.709 & 0.587 & 0.553 & \textbf{0.204} & \textbf{0.006} \\
Milwaukee    &  31 &  15 & 0.500 & 0.057 & 0.057 & 0.519 & \textbf{0.091} & \textbf{0.037} & 0.546 & 0.164 & 0.120 & 0.468 & 0.657 & 0.063 \\
NewYork      &  43 &  36 & 0.569 & 0.085 & 0.113 & 0.532 & 0.192          & 0.171          & 0.538 & 0.214 & 0.151 & 0.559 & 0.116 & 0.171 \\
Oulu         &  66 &  37 & 0.654 & 0.000 & 0.000 & 0.607 & \textbf{0.004} & \textbf{0.148} & 0.526 & 0.214 & 0.332 & 0.496 & 0.518 & 0.517 \\
Oxford       &  10 &  12 & 0.700 & 0.019 & 0.011 & 0.650 & 0.055          & 0.105          & 0.467 & 0.562 & 0.198 & 0.450 & 0.656 & 0.171 \\
Saint Louis  &  17 &  14 & 0.479 & 0.581 & 0.922 & 0.500 & 0.328          & 0.938          & 0.450 & 0.712 & 0.676 & 0.517 & 0.347 & 0.578 \\


\hline
Schizophrenia &  46 &  40 & 0.647 & 0.010 & 0.004 & 0.641 & 0.011 & 0.008 & 0.591 & \textbf{0.076} & \textbf{0.029} & 0.498 & 0.508 & 0.068 \\
\hline
Contex.Disord.&  19 &  19 &   -   &   -   &   -   &   -   &   -   &   -   & 0.757 & 0.002 & 0.010 & 0.558 & \textbf{0.192} & \textbf{0.017} \\

  \end{tabular}
\end{sidewaystable}

\subsection{Simulation: Type I and II error  for Low Sample Size}
\label{sec:simulation}
We conducted a simulation study to quantify the probability of Type I
error and Type II error, for CBT and KTST, in the low sample size
regime, in order to characterize the tests according to the
Neyman-Pearson paradigm (see Section~\ref{sec:np}). Here we expected
that the various sources of variability in the estimation process of
CBT, described in Section~\ref{sec:cbt}, had a negative impact,
increasing the number of errors with respect to that of KTST. The
datasets used are generated according to the description in
Section~\ref{sec:simulated_data}.

In the first part of the simulation, we used the dataset for which
$H_0$ is true ($\delta = 0.0$). Here we quantified $p(\mbox{Type I)}$,
i.e. $H_0$ true but rejected, for CBT and KTST using the DCE embedding
and Gaussian kernel on the sampled graphs. In the same way, in the
second part of the simulation, we used the datasets for which $H_0$ is
false with different degrees of the effect size ($\delta \in \{0.25,
0.5, 0.75, 1.0\})$). With the second part we quantified $p(\mbox{Type
  II})$, i.e. $H_0$ false but not rejected. The results over 1000
repetitions are reported in Table~\ref{tab:simulation}.

\begin{sidewaystable}[t]
  \centering
  \caption{Results of the simulation. The frequency of Type
    I and Type II error (lower is better) of CBT and KTST across
    different effect sizes ($\delta$) are reported, for different
    $p$-value thresholds ($\theta=0.05$ and $\theta=0.01$). In bold are
    indicated the cases in which KTST strongly differ w.r.t. CBT.} 
  \label{tab:simulation}
  \begin{tabular}{ c | c | c | c  c | c  c | c  c | c  c }
    \multicolumn{3}{c}{} & \multicolumn{4}{| c |}{$p(\mbox{Type I})$} & \multicolumn{4}{| c }{$p(\mbox{Type II})$} \\
    \hline
    \multicolumn{3}{c}{} & \multicolumn{2}{| c |}{$\theta = 0.05$} & \multicolumn{2}{| c |}{$\theta = 0.01$} & \multicolumn{2}{| c |}{$\theta = 0.05$} & \multicolumn{2}{| c }{$\theta = 0.01$} \\ 
    \hline
    Ground Truth & $m$ & $n$ & CBT-SVMs & KTST & CBT-SVMs & KTST & CBT-SVMs & KTST & CBT-SVMs & KTST \\
    \hline
    $H_0$ true ($\delta=0$) & 20 & 20 & 0.077 & 0.053 & 0.028 & 0.009 & - & - & - & - \\
    \hline
    $H_0$ false ($\delta=0.25$) & 20 & 20 &  - & - & - & - & 0.813 & 0.781 & 0.907 & 0.919 \\
    $H_0$ false ($\delta=0.5$)  & 20 & 20 &  - & - & - & - & 0.433 & \textbf{0.264} & 0.693 & \textbf{0.498} \\
    $H_0$ false ($\delta=0.75$) & 20 & 20 &  - & - & - & - & 0.093 & \textbf{0.020} & 0.215 & \textbf{0.070} \\
    $H_0$ false ($\delta=1.0$)  & 20 & 20 &  - & - & - & - & 0.013 & 0.000 & 0.018 & 0.001 \\
  \end{tabular}
\end{sidewaystable}

\subsection{Reproducibility of the Results}
All the code used in the experiments and in the simulation study, that
generates the tables and the figures in this section, was developed in
Python using the numerical libraries NumPy and SciPy, together with
the machine learning package
Scikit-learn~\cite{pedregosa2011scikit}. Our code is available under a
Free / OpenSource license
at~\url{https://github.com/emanuele/jstsp2015}.


\section{Discussion}
\label{sec:discussion}

Table~\ref{tab:results_fmri} ad Figure~\ref{fig:p_values_KTST_vs_CBT}
show a strong agreement between the $p$-values of CBT and KTST across
all datasets, supporting the claim of equivalent results between the
two approaches. The Spearman correlation coefficient between all the
54 pairs of $p$-values of CBT and KTST is $0.79$ ($p\mbox{-value} <
0.00001$)\footnote{Estimated with a permutation test and $10000$
  permutations.}, which is highly significant. Notice that this result
holds across multiple datasets from different domains and using both
graph embeddings and graph kernels.

Considering a standard threshold $\theta = 0.05$ to declare the
$p$-value as significant, in Table~\ref{tab:results_fmri} there are
$45$ cases of agreement between CBT and KTST, i.e. when either both
rejected $H_0$ or both did not. The $9$ cases of disagreement,
i.e. when the CBT did not reject $H_0$ while the KTST did (or
viceversa), are reported in Table~\ref{tab:results_fmri} in bold
font. These cases occur for DRE (Berlin, Leipzig, Milwaukee and Oulu),
WL (Schizophrenia) and SP (Beijing, Cambridge, Leipzig, and Contextual
Disorder). For those cases, it is enlightening to see
Figure~\ref{fig:p_values_CBT_DRE_boxplot},
\ref{fig:p_values_CBT_WL_boxplot} and
~\ref{fig:p_values_CBT_SP_boxplot}, where, for each dataset, the great
variability of $acc_{CV}$ is represented in the great variability of
the associated $p$-values, always ranging from significant values to
non-significant ones. From those figures, we can safely conclude that
the disagreement is dominated by variability of CBT and not by
inherent disagreement between CBT and KTST.

In Figure~\ref{fig:p_values_CBT_DCE_boxplot},
\ref{fig:p_values_CBT_DRE_boxplot}, \ref{fig:p_values_CBT_WL_boxplot}
and \ref{fig:p_values_CBT_SP_boxplot} it is shown the great
variability of the $p$-values obtained with $100$ repetitions of the
CBT. This variability is due to the variability of the $acc_{CV}$
estimate, occurring because of the non-deterministic train/test split
during cross-validation and because of the competing joint fit of the
classifier against the estimation of the hyperparameters during model
selection. See Section~\ref{sec:cbt} for the detailed
explanation. Conversely, for KTST, the estimate of $\MMD^2_u$ has no
variability because it is deterministic, given the data. The
variability of its associated $p$-value is due only to the
approximation of the null distribution through permutations. Such
variability can be controlled to the desired level and was negligible
with $M=10000$ permutations in all experiments. The much greater
variability of CBT with respect to KTST is a clear advantage in favor
of the proposed test.

In Section~\ref{sec:simulation}, the effect of the great variability
of CBT in the regime of low sample size is studied with a simulation,
in order to quantify the impact in terms of Type I and Type II error,
in a simplified setting. The results reported in
Table~\ref{tab:simulation} show that when the null hypothesis $H_0$ is
true, i.e. the effect size $\delta = 0$, the rate of false discovery,
i.e. the frequency of Type I error, is almost equivalent for CBT and
KTST, with a marginal $2\%$ advantage in favor of KTST, both when the
significance threshold is $\theta = 0.05$ or $\theta = 0.01$. When
$H_0$ is false, i.e. when $\delta > 0$, the sensitivity of CBT and
KTST, in terms of frequency of Type II error, is computed and reported
in Table~\ref{tab:simulation}. As expected, when the effect size is
too low or too high, i.e. $\delta = 0.25$ and $\delta = 1.0$
respectively, both CBT and KTST behaves similarly. But in the
intermediate cases, i.e. when $\delta = 0.5$ and $\delta = 0.75$, the
advantage of KTST becomes much greater, ranging from $7\%$ to $20\%$
reduction of the Type II error. This results clearly confirms that
KTST is more sensitive than CBT for low sample size, at least in some
cases, and equivalent in other cases. Notice that analogous results
can be obtained when changing the dimension $d$ of the simulated
datasets and after adjusting the effect size $\delta$ accordingly.

As reported in Section~\ref{sec:experiments}, computing the CBT
required hours while KTST required only a few seconds. This is
expected given the description of the amount of the respective
computation in Section~\ref{sec:cbt} and Section~\ref{sec:ktst}, at
least in the setting of our experiments, i.e. low $N$, which is
typical for neuroimaging data. Almost all the time of the computation
is used for the Monte Carlo approximation of the null distribution,
during which the respective test statistics, $acc_{CV}$ and
$\MMD^2_u$, are estimated $M = 10000$ times. Estimating $acc_{CV}$ is
approximately $1000$ times slower that estimating $\MMD^2_u$, because
$acc_{CV}$ is a nested loop, where a classifier is trained and tested
at each iteration. Moreover, in the internal loop, training and
testing is done for each of the hyperparameter values. Notice that
training a classifier usually requires to solve an optimization
problem, so a large number of costly optimizations are
necessary. Conversely, $\MMD^2_u$ requires only $N^2$ sums. This is
another clear advantage of KTST over CBT.





\section{Conclusions}
\label{sec:conclusions}

In this work we proposed the use of the kernel two-sample test (KTST)
for studying systematic differences between two population of graphs
representing brain networks. We compared the KTST with the common use
of classifiers, that here we call classification-based test (CBT). We
claimed that, in general, both tests provide very similar results and,
in Section~\ref{sec:experiments}, we showed it in practice, with
multiple experiments.

We also explained that, for low sample size, the result of CBT may
present high variability \emph{on the same dataset} and gave detailed
description of the causes in Section~\ref{sec:cbt}. This is a major
difference with respect to KTST, which instead has no (or negligible)
variability, given the data. This difference between the two tests is
the motivation of the increased Type II error of CBT with respect
KTST, that we studied with a simulation presented in
Section~\ref{sec:simulation}. A partial remedy to the instability of
the result of CBT is to estimate multiple times $acc_{CV}$, instead of
just once, and to report the median value. This remedy increases the
amount of computations required by the CBT, which is already more than
1000 times greater than those for the KTST, i.e. hours vs. seconds.

One limitation of KTST is the lack of an absolute measure of the
effect size in the data, which means that the KTST can be used to
quantify the significance of a phenomenon but not its effect size. The
$\MMD$ test statistics is not an absolute measure of distance between
distributions, in the sense that, in general, the $\MMD$ value from
two different problems cannot be directly compared. This is different
from the CBT, where performance measures of the classifier, like
accuracy, have an absolute meaning. For this reason, as a final
conclusion, we do not propose to reject the use of classifiers in the
domain of brain networks, but to use both KTST and classifiers to
study significance and effect size, respectively.

As future work, we plan to investigate the characteristic property of
graph kernels, which, up to now, has never been proved. As mentioned
in Section~\ref{sec:ktst}, a \emph{consistent} KTST requires a
characteristic graph kernel. Anyway, this is not very different from
using a kernel based on graph embedding plus the Gaussian kernel, as
we do in this study. Even though the Gaussian kernel is
characteristic, no results are known about mixing it with graph
embeddings. Notice that, to the best of our knowledge, also the
consistency of the CBT has never been proved.



\end{document}